\documentclass[10pt,twocolumn,a4paper]{article}

\usepackage[final]{cvww}

\usepackage[pagebackref,breaklinks,colorlinks,allcolors=cvwwblue]{hyperref}

\usepackage[nolist,nohyperlinks]{acronym}

\usepackage{multirow}
\usepackage{array}
\usepackage{lineno}
\newcolumntype{F}{>{\footnotesize}r}   % small left-aligned
\newcolumntype{S}{>{\small}r}   % small left-aligned
\def\paperID{19}

\title{Koo-Fu CLIP: Closed-Form Adaptation of Vision–Language Models via Fukunaga–Koontz Linear Discriminant Analysis}

\author{Matej Suchanek, Klara Janouskova, Ondrej Vasatko, Jiri Matas
\vspace{2mm}\\
Visual Recognition Group\\
Department of Cybernetics\\
Faculty of Electrical Engineering\\
Czech Technical University in Prague\\
}

\begin{document}
\maketitle

\newcommand{\knn}{\textit{k}-nearest neighbors}
\newcommand{\imnet}[1][1]{ImageNet-#1K}
\newcommand{\kf}{Koo-Fu CLIP}
\newcommand{\real}{ReaL}

\begin{acronym}
\acro{llm}[LLM]{Large Language Model}
\acro{cnn}[CNN]{Convolutional Neural Network}
\acro{clip}[CLIP]{Contrastive Language–Image Pre-training}
\acro{mllm}[MLLM]{Multimodal Large Language Model}
\acro{vlm}[VLM]{Vision-Language Model}
\acro{lda}[LDA]{Linear Discriminant Analysis}
\acro{mda}[MDA]{Multiple Discriminant Analysis}
\acro{nvp}[NVP]{nearest visual prototype}
\acro{knn}[\textit{k}-NN]{\knn{}}
\end{acronym}

\newcommand{\ondra}{\textcolor{blue}{[Ondra]}}
\newcommand{\matej}{\textcolor{green}{[Matej]}}

\newcommand{\red}[1]{{\color{red}#1}}
\newcommand{\todo}[1]{{\color{red}#1}}
\newcommand{\TODO}[1]{\textbf{\color{red}[TODO: #1]}}

%%%%%%%%% BODY TEXT
% Best practice is to split the contents of your paper into separate .tex
% files, as shown below:

\begin{abstract}
Visual–language models such as CLIP provide powerful general-purpose representations, but their raw embeddings are not optimized for supervised classification, often exhibiting limited class separation and excessive dimensionality. We propose Koo-Fu CLIP, a supervised CLIP adaptation method based on \textbf{Fu}kunaga–\textbf{Koo}ntz Linear Discriminant Analysis, which operates in a whitened embedding space to suppress within-class variation and enhance between-class discrimination. The resulting closed-form linear projection reshapes the geometry of CLIP embeddings, improving class separability while performing effective dimensionality reduction, and provides a lightweight and efficient adaptation of CLIP representations.

Across large-scale ImageNet benchmarks, nearest visual prototype classification in the Koo-Fu CLIP space improves top-1 accuracy from 75.1\% to 79.1\% on ImageNet-1K, with consistent gains persisting as the label space expands to 14K and 21K classes. The method supports substantial compression by up to 10–12× with little or no loss in accuracy, enabling efficient large-scale classification and retrieval.
\end{abstract}

\section{Introduction}
\label{sec:intro}
% Visual language models, VLMs in short,  have had an enormous impact on computer vision .... e.g. zero-short recognition, .....
Visual-language models (VLMs) have fundamentally changed computer vision by learning joint image–text representations in a shared embedding space that aligns visual and linguistic concepts, unlocking capabilities such as zero-shot recognition,  open-vocabulary classification, and cross-modal retrieval.

% CLIP is a canonical case, it is the original VLM. Something about CLIP - learning, data. It has been applied to a wide range of problems: ..... i treba deep fake detection \cite{Yermakov-WACV2026}
\ac{clip} \cite{radford2021learning} is the canonical example of a VLM, trained via large-scale contrastive learning on image–text pairs, and its strong, general-purpose embeddings have made it a foundational model for a wide range of vision tasks.
Beyond standard classification, CLIP embeddings have been successfully applied to diverse problems, including open-set recognition and out-of-distribution detection \cite{jiang2024negative,miyai2025zero}, domain adaptation \cite{addepalli2024leveraging}, retrieval \cite{radford2021learning}, 
% robustness analysis,
and security-related tasks such as deepfake detection \cite{ojha2023towards,yermakov2025unlocking}.

% Many CLIP modifications and improvements have been proposed, SIGLIP, .....
% TODO expand with applications

% a bit on limitations
Despite their versatility, raw CLIP embeddings are not optimized for downstream supervised classification, often exhibiting suboptimal class separation and unnecessarily high dimensionality when applied to fixed label spaces such as \imnet[1] or \imnet[21]. Numerous methods have been proposed to adapt CLIP to downstream tasks, including prompt learning (\eg, CoOp \cite{coop} and CoCoOp \cite{cocoop}), cache-based methods (\eg, Tip-Adapter \cite{zhang2021tip}), and linear probing \cite{chen2020simple,radford2021learning}, each trading off adaptation strength, computational cost, and data efficiency.

In this paper, we propose \kf{}, a supervised CLIP adaptation method based on \textbf{Fu}kunaga-\textbf{Koo}ntz \ac{lda}. The Fukunaga-Koontz transformation applies \ac{lda} in a whitened feature space to jointly remove within-class correlations and maximize between-class scatter. 
This yields a closed-form linear projection that reshapes the geometry of CLIP embeddings, improving class separability and enabling principled low-dimensional discriminative subspaces.
\kf{} bridges contrastive representation geometry, classical discriminative subspace learning, and lightweight adaptation of foundation models.

\Ac{nvp} classification in the \kf{} visual embedding space substantially improves over the original embeddings. On \imnet[1], learning the transform from \imnet[1] training samples improves top-1 accuracy from 75.1\% to 79.1\%, with comparable gains observed under both standard and re-annotated evaluation protocols.
Importantly, these improvements remain consistent as the number of candidate classes increases, persisting when extending the label space to 14K and 21K classes, which introduces a large number of distractors and closely resembles an image retrieval scenario.
This robustness suggests that \kf{} effectively sharpens class discrimination rather than overfitting to a closed-set regime. 

In addition, the method enables substantial dimensionality reduction with minimal loss of accuracy: 
reducing the embedding dimensionality by a factor of 3× (768 → 256), incurs less than 0.5\% degradation in \ac{nvp} performance, while simultaneously yielding a +1\% gain for 10-NN classification.
Even more aggressive reductions of 10–12× (76–64 dimensions) retain accuracy on par with or better than the original CLIP space. While \knn{} classification in the transformed space provides a modest 1–2\% accuracy advantage over \ac{nvp}, it requires storing and searching over all training embeddings, resulting in roughly 1000× higher memory and computational cost.
In contrast, \ac{nvp} achieves competitive performance using only class prototypes, highlighting \kf{} as a particularly attractive solution for efficient large-scale classification and retrieval.

% if we improve k-nn, we can expect improvements for retrieval? --> eval for ICML/ICCV

The contributions of this paper are:
\begin{itemize}
    \item A novel closed-form adaptation of CLIP embeddings via discriminative whitening, achieving consistent accuracy gains across vision-only and zero-shot evaluation paradigms.
    \item Effective dimensionality reduction (up to 10--12×) with minimal accuracy loss.
    \item Extensive ablation and analysis of the effects of regularization, dimensionality reduction, neighbor count, distance metrics, and evaluation protocols, providing practical guidance for deployment.
\end{itemize}
\section{Related Work}
\label{sec:rw}
The proposed \kf{} lies at the intersection of contrastive representation geometry, supervised discriminative subspace learning, and efficient adaptation of foundation models.

\noindent\textbf{CLIP and its extensions.}
Vision–language models such as CLIP~\cite{radford2021learning} learn joint image–text embeddings through large-scale contrastive pretraining,
enabling strong zero-shot and transfer performance across a wide range of visual recognition tasks.
Subsequent models, including SigLIP~\cite{zhai2023sigmoid} and SigLIP2~\cite{tschannen2025siglip}, refine the training objective and scalability of this paradigm
while preserving the core idea of a shared embedding space.
Although these models produce highly general-purpose representations, 
downstream adaptation is typically performed via linear classifiers, 
prompt optimization, 
or lightweight fine-tuning,
leaving the underlying embedding geometry unchanged.

Owing to their contrastive training objective, 
CLIP-style models organize visual representations in a highly structured geometric space,
where semantic similarity is reflected by angular proximity and class information is distributed across many directions.
Recent analyses interpret such representations through mixture-based or Fisher-style frameworks~\cite{bansal2024understanding}, 
suggesting that a significant fraction of the embedding space is redundant for a given downstream classification task~\cite{chen2023wdiscood}.
% As a result, CLIP embeddings are high-dimensional and expressive, yet not explicitly optimized for supervised class separation.

% Building on these observations,
% \kf{} leverages supervised discriminative structure to reshape CLIP embeddings into a compact subspace optimized for class separation.

\noindent\textbf{CLIP adaptation and dimensionality reduction.}
Most existing CLIP adaptation methods operate at the level of classifiers, prompts, or similarity aggregation \cite{zhang2021tip}.
Linear probing evaluates frozen representations using a learned linear classifier,
while adapter-based approaches such as Tip-Adapter~\cite{zhang2021tip} refine decision boundaries through feature caches.
Prompt learning methods, including CoOp, CoCoOp, and inverse prompting~\cite{coop,cocoop,inverse_prompts}, adapt the textual side of CLIP via optimization.
While effective, these approaches preserve the original embedding space and typically require iterative training.

In contrast, \kf{} performs a closed-form, supervised transformation of the visual embeddings themselves.
By applying LDA, we obtain a compact discriminative subspace that improves class separability while enabling principled dimensionality reduction, without backpropagation or additional inference-time components.

\noindent\textbf{LDA and discriminative subspaces in deep representations.}
Linear Discriminant Analysis has a long history as a supervised dimensionality reduction technique,
but its use in modern deep representations has been relatively limited.
A notable computer vision exception is WDiscOOD~\cite{chen2023wdiscood} which exploits whitened LDA subspaces of CLIP embeddings for confidence estimation and out-of-distribution detection.
In contrast, we demonstrate that the same discriminative structure can be directly leveraged for representation-level adaptation, yielding significant improvements in classification accuracy while compressing CLIP representations.
Our work thus repositions LDA not as an auxiliary analysis tool, but as an effective and scalable adaptation mechanism for foundation model embeddings.

\noindent\textbf{ImageNet}
\imnet[21]~\cite{deng2009imagenet} is a large-scale visual recognition dataset organized according to the WordNet hierarchy \cite{miller1995wordnet}, comprising over 21,000 object categories with substantial semantic granularity and class imbalance.
\imnet[1]~\cite{russakovsky2015imagenet} was later derived as a curated subset of this hierarchy, focusing on 1,000 commonly occurring categories and augmented with fixed training, validation, and test splits to support standardized benchmarking.

Despite its known limitations, \imnet[1] remains the de facto benchmark for large-scale image classification and representation learning, serving as a common reference point across the literature~\cite{rw2019timm}.
To address known issues in the original \imnet[1] annotations~\cite{kisel2025flaws,reannot_context,reannot_eval,reannot_pervasive,reannot_arewedone}, including label noise, semantic overlap, and images containing multiple valid labels, numerous reannotation efforts have been proposed.
In this work, we follow prior practice and evaluate on a cleaner validation protocol~\cite{kisel2025flaws} obtained by taking the union of these correction efforts.

\section{Method}
\label{sec:method}
This section first overviews the theoretical background of the Fukunaga-Koontz transformation and then explains how its applied to the CLIP model. Finally, we outline the CLIP-based classifiers used in our experiments.
\subsection{Preliminary}
\label{sec:method:prelim}

\paragraph{Linear Discriminant Analysis}
Linear Discriminant Analysis (LDA) is a supervised statistical method used to reduce the dimensionality of a feature space while preserving its discriminative capabilities~\cite{fisher1936use}. Its advantages include simplicity and efficiency, since no training data need to be accessed during inference, making it applicable even for learning from datasets with millions of data points.
% Furthermore, ordinary classification techniques, such as \emph{k-nearest neighbors}, can be used in the reduced feature space.
%, which improves their efficiency.

\ac{lda} finds a linear mapping from the original feature space to a lower-dimensional subspace in which the class-specific mean vectors are maximally separated, while the variance (scatter) of samples within each class is minimized.
Formally, given a labeled dataset $\{(\mathbf{x}_i, y_i)\}_{i=1}^N$ with samples $\mathbf{x}_i \in \mathbb{R}^D$ and class labels $y_i \in \{1,\dots,K\}$, let
\begin{equation}
    {\boldsymbol{\mu}} = \frac{1}{N} \sum_{i} \mathbf{x}_i
\end{equation}
be the global data mean vector,
\begin{equation}
    \boldsymbol{\mu}_k = \frac{1}{N_k} \sum_{i: y_i=k} \mathbf{x}_i
\end{equation}
per-class mean vectors, where $N_k$ is the number of samples in class $k$,
\begin{equation}
    \mathbf{S}_k = \sum_{i: y_i=k} (\mathbf{x}_i - \boldsymbol{\mu}_k)(\mathbf{x}_i - \boldsymbol{\mu}_k)^\top
\end{equation}
conditional scatter matrix of class $k$, and
\begin{align}
    \mathbf{S}_w &= \sum_{k=1}^K \mathbf{S}_k, \\
    \mathbf{S}_b &= \sum_{k=1}^K N_k ({\boldsymbol{\mu}}_k - \boldsymbol{\mu}) ({\boldsymbol{\mu}}_k - \boldsymbol{\mu})^\top, \label{eq:S_b}
\end{align}
within-class and between-class scatter matrices. \ac{lda} aims to find a projection matrix $\mathbf{W} \in \mathbb{R}^{D \times L}$ such that the objective \begin{equation}
%\mathbf{W}^* = \arg\max_{\mathbf{W} \in \mathbb{R}^{d \times k}}
\frac{\det(\mathbf{W}^\top \mathbf{S}_b \mathbf{W})}{\det(\mathbf{W}^\top \mathbf{S}_w \mathbf{W})}
\end{equation}
is maximal. The solution can be found in closed form by solving the generalized eigenvalue problem
\begin{equation}
    \mathbf{S}_b \mathbf{v} = \lambda \mathbf{S}_w \mathbf{v}.
\end{equation}

Finally, to perform the mapping to a lower-dimensional space, the eigenvalues $\lambda_j$ are sorted in descending order ($\lambda_1 \geq \lambda_2 \geq \dots \geq \lambda_D$) and the ultimate projection matrix, defining a mapping into an $L$-dimensional subspace, is formed by the top $L$ eigenvectors $\mathbf{v}_j$ corresponding to the highest eigenvalues:
\begin{equation}
    \mathbf{W}_L = [\mathbf{v}_1, \dots, \mathbf{v}_L].
\end{equation}
Note that due to the inherent singularity of the matrix $\mathbf{S}_b$, the number of eigenvectors produced is always lower than the number of classes $K$. %\cite{bishop2006pattern}

\paragraph{Fukunaga-Koontz Transform}

\ac{lda} can be improved by the so-called Fukunaga-Koontz Transform \cite{fukunaga1990introduction}.
This method proposes several enhancements over the standard \ac{lda}, posing its generalization.
In particular, it introduces the so-called \emph{whitening transformation} followed by a \emph{discriminative rotation}.
This is done by simultaneous diagonalization of the within-class and between-class scatter matrices, thereby normalizing within-class variance, achieving greater class separation, and overcoming the inherent limitation on the number of produced directions.

The whitening begins with the eigendecomposition of the within-class scatter matrix:
\begin{equation}
    \mathbf{S}_w = \mathbf{V} \mathbf{\Lambda} \mathbf{V}^\top. \label{eq:S_w_eigen}
\end{equation}
From this, we compute the inverse square root matrix
\begin{equation}
    \mathbf{S}^{-1/2}_w = \mathbf{V} \mathbf{\Lambda}^{-1/2} \mathbf{V}^\top,
\end{equation}
where $\mathbf{\Lambda}^{-1/2}$ is the diagonal matrix of the inverse square roots of the eigenvalues ($\sqrt{\frac{1}{\lambda_i}}$). Satisfying
\begin{equation}
     \mathbf{S}^{-1/2}_w \mathbf{S}^{}_w \mathbf{S}^{-1/2}_w = \mathbf{I},
\end{equation}
the matrix $\mathbf{S}^{-1/2}_w$ effectively performs the diagonalization of the within-class scatter matrix, transforming the feature space to another where per-class variances become spherical.

% \paragraph{Whitening Transformation}
% We decouple \ac{lda} into a whitening step followed by a rotation~\cite{fukunaga1990introduction}. First, we perform the eigendecomposition of the regularized within-class scatter $\hat{\mathbf{S}}_w = \mathbf{V} \mathbf{\Lambda} \mathbf{V}^\top$. We then compute the whitening transformation matrix $\mathbf{W}$ via the inverse square root:
% \begin{equation}
%     \mathbf{W} = \mathbf{V} \mathbf{\Lambda}^{-1/2} \mathbf{V}^\top.
% \end{equation}
% Applying this transformation effectively spheres the intra-class distributions, ensuring that the residual variance is isotropic and unitary in all directions~\cite{lee2018simple}. \todo{Is this citation relevant? Which exact part of the paper is cited?}

We note that in order to compute the inverse square root, all eigenvalues need to be positive. In our experiments, however, we encounter $\mathbf{S}_w$ matrices with negative eigenvalues, even though they are positive semi-definite by definition. We attribute this to the fact that in high-dimensional feature spaces or scenarios with limited samples per class, the empirical $\mathbf{S}_w$ often becomes ill-conditioned or singular~\cite{friedman1989regularized}. As a mitigation, we apply shrinkage regularization parametrized by $\lambda$ prior to whitening \cite{ledoit2004well}:
\begin{align}
    \mathbf{S}'_w =&\ \mathbf{S}^{}_w + \lambda \mathbf{I}, \label{eq:S_w_reg} \\
    \mathbf{Z} :=&\ \mathbf{S}^{-1/2}_w.
\end{align}
This parameter $\lambda$ is the only hyper-parameter of the proposed method.

We formally define the whitening function $f(\cdot)$ as the projection of a centered vector onto the sphered space:
\begin{equation}
    f(\mathbf{x}) = \mathbf{Z} \mathbf{x}.
\end{equation}

Then, the between-class scatter matrix (\cref{eq:S_b}) is computed in the whitened space, which is defined as the range of the matrix $\mathbf{S}^{-1/2}_w$. This requires mapping the global and per-class means to the whitened space:
\begin{equation}
    \mathbf{S}'_b = \sum_{k=1}^K N_k f(\boldsymbol{\mu}_k - \boldsymbol{\mu}) f(\boldsymbol{\mu}_k - \boldsymbol{\mu})^\top. \label{eq:Sb-after-whitening}
\end{equation}
Next, we perform the eigendecomposition $\mathbf{S}'_b = \mathbf{U} \mathbf{\Gamma} \mathbf{U}^\top$, where the columns of $\mathbf{U}$ represent the directions of maximum class separability.
% The projection matrix is constructed by chaining the whitening transform and the discriminative rotation:
% \begin{equation}
%     \mathbf{T} = \mathbf{U}^\top \mathbf{W}.
% \end{equation}
Similarly to \ac{lda}, dimensionality reduction is performed by sorting the eigenvalues of the whitened between-class scatter matrix $\mathbf{S}'_b$ ($\gamma_1 \geq \gamma_2 \geq \dots \geq \gamma_D$) and truncating the rotation matrix $\mathbf{U}$ to keep only the eigenvectors corresponding to the top $L$ eigenvalues:
\begin{align}
    \mathbf{U}_L = [\mathbf{u}_1, \dots, \mathbf{u}_L]
\end{align}
%Components associated with smaller eigenvalues (\ie, where $\gamma_j \approx 0$) typically correspond to noise or non-discriminative variations and are discarded.

The ultimate projection matrix $\mathbf{T} \in \mathbb{R}^{L \times D}$ is obtained by composing the truncated rotation with the whitening transformation:
\begin{equation}
    \mathbf{T} = \mathbf{U}_L^\top \mathbf{Z}. \label{eq:final}
\end{equation}
This projection maps the original features into an optimized $L$-dimensional discriminative subspace.

% To construct the final compact feature space, we utilize the spectral properties of the between-class scatter matrix. The eigenvalues $\gamma_j$ derived from the decomposition $\tilde{\mathbf{S}}_b = \mathbf{U} \mathbf{\Gamma} \mathbf{U}^\top$ represent the discriminative power of the corresponding eigenvectors $\mathbf{u}_j$~\cite{bishop2006pattern}. We sort the eigenvalues in descending order such that $\gamma_1 \geq \gamma_2 \geq \dots \geq \gamma_D$. To reduce the dimensionality from $D$ to $M$, we select the top $M$ eigenvectors corresponding to the highest eigenvalues, forming the truncated rotation matrix $\mathbf{U}_L = [\mathbf{u}_1, \dots, \mathbf{u}_L]$. Components associated with smaller eigenvalues (where $\gamma_j \approx 0$) typically correspond to noise or non-discriminative variations and are discarded.

\subsection{Koo-Fu CLIP}
\label{sec:pipeline}

Our proposed method {\bf \kf{}} applies the Fukunaga-Koontz Transform to embedding spaces of CLIP-based models.
These models serve as image encoders -- they turn images into compact embeddings (vectors) from a feature space where visual and semantic similarity can be expressed mathematically.
The core idea of our method is to augment the encoding process with a transformation that adapts the implicit feature space to a downstream task and improves the accuracy of common classification techniques.
This transformation is learned in a supervised manner, from a manually annotated dataset.

The overall approach can be summarized in the following pipeline:
\begin{enumerate}
    \item Use a CLIP-based to extract embeddings of training images.
    \item Compute the necessary statistics as described in \Cref{sec:method:prelim}.
    \item Perform (cross-)validation to select the best transformation model for a given classification method. For example, the $\lambda$ regularization parameter and the number of preserved dimensions can be ablated.
    \item At inference time, map the extracted embeddings in the optimized feature space (\cref{eq:final}) and then perform classification.
\end{enumerate}

\subsection{Classifiers}

We also present classification methods that are commonly applied in high-dimensional feature spaces of CLIP-based models.
% \todo{We evaluate classification techniques that maintain computational efficiency and high precision within the high-dimensional feature spaces typical of CLIP-based models. In the following sections, we outline the principles of these methods and describe their adaptation to transformed CLIP embeddings to facilitate the empirical comparisons presented in our results. }

\paragraph{Classification by \ac{knn}}
This algorithm keeps a database of all training embeddings.
During inference, a sample is classified by retrieving the $k$~nearest training samples and choosing the most common label among them.
The optimal value for~$k$ is determined by (cross-)validation.
A limitation of this well-established approach is the requirement to store the entire database in memory, with increased memory overhead traded for speed unless inexact answers are acceptable.

\paragraph{Classification by nearest visual prototype (\ac{nvp})}
Before inference, the training images are first accumulated in class-wise \emph{prototypes}.
This is usually the mean embedding vector (or weighted mean) of all samples in each class.
At inference time, samples are classified by the nearest prototype.
As the number of prototypes is usually marginal compared to the number of training samples, which all potentially need to be accessed for \ac{knn}, this approach is more memory efficient.

Furthermore, it also enables smoother advancement from closed-set to open-set classification tasks, as it is possible to amend the prototype space with a new prototype originating from a few representative samples of a concept that was not present in the training data. This is also referred to as ``few-shot classification''.

\paragraph{Zero-shot classification}
This technique leverages the capabilities of CLIP-based models to retrieve images using queries in natural language.
These models are trained to embed both images and text queries into a common feature space in a contrastive manner -- text embeddings should be similar to embeddings of their visual representations \cite{radford2021learning}.
This eliminates the need for gathering sufficient number of image instances for a given concept (hence ``zero-shot''); instead, the prototype is found directly by encoding the label (name) of the concept, which allows advancing from closed-set to open-set classification tasks.

However, since CLIP-based models are often trained on images with captions scraped from the internet \cite{radford2021learning}, search queries (called \emph{prompts}) consisting only of the name might not match the training distribution (\eg, an image is more likely to be described as ``a photo of an apple'' rather than just ``apple'').
Therefore,
%and based on \todo{previous/recent} work \todo{[cite]},
our experiments also discuss the influence of different prompt strategies and templates.

\section{Experiments}
\label{sec:experiments}

% To demonstrate the benefits of our method, we conduct a series of experiments with various preconditions, such as different input data, hyper-parametrization, and evaluation protocols.

\subsection{Setup}
\label{sec:setup}

\noindent\textbf{Model}
We choose OpenAI's original pretrained CLIP \mbox{ViT-L/14} model \cite{radford2021learning} as the VLM for our experiments, which produces 768-dimensional embeddings. %\todo{Don't forget to update this in case we included SIGLIP \cite{tschannen2025siglip}.}

\noindent\textbf{Datasets}
We use \textit{\imnet[21]} \cite{deng2009imagenet} (Winter 2021 edition, abbr. \mbox{IN-21k}), a large-scale dataset of images grouped by the concept they depict.
This revision contains over 13,150,000 images categorized into 19,167 classes.

In addition, we use \textit{\imnet[1]} \cite{russakovsky2015imagenet} (abbr. \mbox{IN-1k}), a subset of \imnet[21], which contains over 1,280,000 images categorized into (in most cases) evenly distributed 1,000 classes.
Unlike \imnet[21], this subset also comes with an independent validation set of 50,000 images.
Except for one, the classes of \mbox{IN-1k} are contained in \mbox{IN-21k}.
We also evaluate on a dedicated subset of \mbox{IN-21k} classes, denoted 14K, corresponding to 14,271~leaves in the ImageNet class tree.
See \cref{tab:synset_groups} for an overview of ImageNet class sets and their overlaps.

Recently, a significant effort has been invested to assess and improve the quality of \mbox{IN-1k} annotations \cite{kisel2025flaws,reannot_context,reannot_eval,reannot_pervasive,reannot_arewedone}.
Accordingly, we also evaluate our methods on a label-corrected \textit{“cleaner” validation} subset of more than 36,000 images combined from multiple re-annotation initiatives \cite{kisel2025flaws}. Note that this subset may inflate performance due to its bias towards easier samples (it excludes images where reannotations disagree) and a more permissive evaluation metric (see next paragraph).

\begin{table}[t]
    \centering
    \setlength{\tabcolsep}{4pt}
    \begin{tabular}{ c|c|cllr }
        \toprule
        \multicolumn{3}{c}{Abbr.} & \multicolumn{1}{c}{Description} & \multicolumn{2}{r}{No. of classes} \\
        \midrule
        \multicolumn{1}{c}{} & \multicolumn{1}{c|}{\multirow{3}{*}{14K}} & \multicolumn{1}{c}{\multirow{2}{*}{1K}} & \multicolumn{2}{l}{in 1k, not in 21k} & 1 \\
        \cline{1-1}
        \multirow{5}{*}{21K} & & & \multicolumn{2}{l}{in 1k, and in 21k} & 999 \\
        \cline{3-3}
        & & \multicolumn{1}{c}{} & \multicolumn{2}{l}{leaves outside 1k} & 13,271 \\
        \cline{2-2}
        & \multicolumn{2}{c}{} & \multicolumn{2}{l}{ancestors of 1k outside 1k}   &    757 \\
        & \multicolumn{2}{c}{} & \multicolumn{2}{l}{descendants of 1k outside 1k} &  1,234 \\
        & \multicolumn{2}{c}{} & \multicolumn{2}{l}{other} &  2,906 \\
        \midrule
        \multicolumn{6}{r}{\textbf{Total: 19,167 + 1}} \\
        \bottomrule
    \end{tabular}
    \caption{
    ImageNet classes (WordNet synsets \cite{miller1995wordnet}) categorized by their relation to \imnet[1] and their position within the WordNet hierarchy.
    The first two rows form the \imnet[1] classes. 
    Their union with the non-1k leaf nodes, denoted as 14K in experiments, consists of 14,271 classes with no descendant-ancestor relationships; these sets are mutually exclusive in images depicting a single ImageNet object.
    \imnet[21] contains approximately 19,000 classes, following the removal of problematic synsets in a revision \cite{ImageNet2019Update}.
    % Interestingly, the \imnet[1] classes have on average significantly more ancestral classes than the remaining sets in IN-21k.
    % While \imnet[1] is not hierarchical, specific sets within the hierarchy are fully contained within their respective complement in \imnet[21].
    }
    \label{tab:synset_groups}
\end{table}

\begin{table}[t]
\setlength{\tabcolsep}{4pt}
% \small
    \centering
    \begin{tabular}{F rrr rrrr}
        \toprule
        & \multicolumn{1}{c}{Space} & & \multicolumn{2}{c}{Top-1} & \multicolumn{2}{c}{Top-5} \\
        \# &
        \multicolumn{1}{c}{$f(\cdot)$} &
        \multicolumn{1}{c}{class.} &
        \multicolumn{1}{c}{Clean} &
        \multicolumn{1}{c}{Val.} &
        \multicolumn{1}{c}{Clean} &
        \multicolumn{1}{c}{Val.} \\
        \midrule
        \textcolor{gray}{1} &  (--) & \multirow{3}{*}{1K}  & 85.82 & 75.08 & 96.84 & 93.42 \\
        \textcolor{gray}{2} &  (1K) &               & \textbf{89.71} & \textbf{79.05} & \textbf{98.19} & \textbf{95.54} \\
        \textcolor{gray}{3} & (21K) &               & 88.11 & 78.40 & 98.03 & 95.31 \\
        \cmidrule{2-7}
        \textcolor{gray}{4} &  (--) & \multirow{3}{*}{14K}  & 65.76 & 57.79 & 73.27 & 70.03 \\
        \textcolor{gray}{5} &  (1K) &  & \textbf{68.28} & \textbf{60.52} & \textbf{74.15} & \textbf{71.39} \\
        \textcolor{gray}{6} & (21K) &  & 67.96 & 60.09 & 74.08 & 71.23 \\
        \cmidrule{2-7}
        \textcolor{gray}{7} &  (--) & \multirow{3}{*}{21K} & 50.96 & 42.73 & 80.15 & 71.28 \\
        \textcolor{gray}{8} &  (1K) &               &\textbf{ 55.37} & \textbf{46.84} & \textbf{84.25} & \textbf{75.70} \\
        \textcolor{gray}{9} & (21K) &               & 53.71 & 45.32 & 83.41 & 74.76 \\
        \bottomrule
    \end{tabular}
    \caption{
    {\bf Accuracy of nearest visual prototype classification (\ac{nvp})}, with the class mean embedding the single prototype, in the CLIP embedding space: unmodified \mbox{(--)} and transformed by $f(\cdot)$; the inverse square root of the within-class scatter matrix $\mathbf{S}_{w:\mathrm{1K}}$ computed on the IN-1k training set (1K), and 
    $\mathbf{S}_{w:\mathrm{21K}}$, computed on the full IN-21k (21K) respectively. \\
    Accuracy is reported for the IN-1k Validation 
    % (Val. column, comparable with published results)
    and the label-corrected ``Cleaner'' validation sets.
    % which approximates true performance well as it has many IN-1k labelling errors corrected.
    % \\
    Performance is evaluated on the standard \imnet[1] close-set recognition task with 1k classes, as well as in setups with distractors. For details about the 1K, 14K, and 21K classes, see \cref{tab:synset_groups}.
    }
    \label{tab:vnp_clip}
\end{table}
\begin{table}[htb]
    \setlength{\tabcolsep}{3pt}
    \centering
    \begin{tabular}{F r r c cccccc}
        \toprule
        & Space & & \multicolumn{2}{c}{1-NN} & \multicolumn{2}{c}{10-NN} & \multicolumn{2}{c}{15-NN} \\
        \# &
        \multicolumn{1}{c}{$f(\cdot)$} &
        \multicolumn{1}{c}{class.} &
        \multicolumn{1}{c}{Clean} &
        \multicolumn{1}{c}{Val.} &
        \multicolumn{1}{c}{Clean} &
        \multicolumn{1}{c}{Val.} &
        \multicolumn{1}{c}{Clean} &
        \multicolumn{1}{c}{Val.} \\
        \midrule
        \textcolor{gray}{1} & (--) & \multirow{3}{*}{1K} & 84.9 & 76.3 & 88.7 & 79.6 & 88.7 & 79.5 \\
        \textcolor{gray}{2} & (1K) &  & \textbf{86.1} & \textbf{77.6} & \textbf{89.9} & \textbf{81.0} & \textbf{90.1} & \textbf{81.0} \\
        \textcolor{gray}{3} & (21K) &  & 85.7 & 77.3 & 89.6 & 80.7 & 89.8 & 80.7 \\
        \cmidrule{2-9}
        \textcolor{gray}{4} & (--) & \multirow{3}{*}{14K} & 59.2 & 51.2 & 68.0 & 58.7 & 68.5 & 59.2 \\
        \textcolor{gray}{5} & (1K) &  & \textbf{60.9} & \textbf{52.8} & \textbf{69.8} & \textbf{60.4} & \textbf{70.5} & \textbf{61.0} \\
        \textcolor{gray}{6} & (21K) &  & 60.4 & 52.3 & 69.2 & 59.8 & 69.9 & 60.4 \\
        \cmidrule{2-9}
        \textcolor{gray}{7} & (--) & \multirow{3}{*}{21K} & 47.4 & 40.8 & 57.5 & 49.5 & 58.4 & 50.2 \\
        \textcolor{gray}{8} & (1K) &  & \textbf{48.8} & \textbf{42.2} & \textbf{59.0} & \textbf{50.9} & \textbf{60.2} & \textbf{51.9} \\
        \textcolor{gray}{9} & (21K) &  & 48.3 & 41.7 & 58.3 & 50.2 & 59.5 & 51.3 \\
        \bottomrule
    \end{tabular}

    \caption{
        {\bf Accuracy of \knn{} (\ac{knn})}, with the class mean embedding the single prototype, in the CLIP embedding space.
        The notation and settings are analogous to \cref{tab:vnp_clip}.
        Results for classification by 1-, 10- and 15-nearest neighbors are reported instead of \mbox{top-5} accuracy, all corresponding to the \mbox{Top-1} columns in \cref{tab:vnp_clip}.
    }
    \label{tab:knn_cosine_clip}
\end{table}

\begin{table}[h]
    \setlength{\tabcolsep}{3pt}
    \centering
    \begin{tabular}{crrrrrrr}
        \toprule
        Space & & 
        \multicolumn{2}{c}{{\it synset}} 
          & \multicolumn{2}{c}{{\it synset}$_7$} 
          & \multicolumn{2}{c}{{\it mod}$_7$} \\
        $f(\cdot)$ &
        \multicolumn{1}{c}{$\lambda$} &
        \multicolumn{1}{c}{Clean} &
        \multicolumn{1}{c}{Val.} &
        \multicolumn{1}{c}{Clean} &
        \multicolumn{1}{c}{Val.} &
        \multicolumn{1}{c}{Clean} &
        \multicolumn{1}{c}{Val.} \\
        \midrule
        (--) &  --- & 79.60 & 69.79 & 83.42 & 73.50 & 85.53 & 75.38 \\
        \midrule
        (1K) &   50 &  1.09 &  0.90 &  1.96 &  1.62 &  4.45 &  3.79 \\
        (1K) &  150 & 11.93 & 10.06 & 26.61 & 22.34 & 45.34 & 38.55 \\
        (1K) &   1k & 71.89 & 62.08 & 80.58 & 70.73 & 83.29 & 73.36 \\
        (1K) &   3k & 78.42 & 68.52 & 83.50 & 73.56 & 85.51 & 75.47 \\
        (1K) &   5k & 79.37 & 69.54 & 83.77 & 73.81 & 85.79 & 75.76 \\
        (1K) &  10k & \textbf{79.87} & \textbf{70.02} & \textbf{83.87} & \textbf{73.96} & \textbf{85.87} & \textbf{75.85} \\
        (1K) & 100k & 79.72 & 69.92 & 83.52 & 73.63 & 82.92 & 72.65 \\
        %(21K) &   250 &  1K &  0.48 &  0.41 &  2.09 &  1.79 \\
        %default &   --- & 21K & 41.35 & 35.04 & & \\
        %     1K &    50 & 21K &  0.13 &  0.09 & & \\
        %    21K &   250 & 21K &  0.00 &  0.00 & & \\
        \bottomrule
    \end{tabular}
    \caption{
        Zero-shot classification, \ie, textual nearest prototype (tNP), in the CLIP space with gradually regularized whitening with $\mathbf{S}'_{w:\mathrm{1K}}$. \mbox{Top-1} accuracy evaluated on the IN-1k Clean and validation sets with the following prompts: \\
        \textit{synset} -- ``image of \{class name\}'', \\
        \textit{synset}$_7$ -- mean of 7 prompt embeddings of templates from \cite{kisel2025flaws}, \\
        \textit{mod}$_7$ -- mean of the 7 prompt templates, where IN-1k class names are replaced with OpenAI optimized names \cite{radford2021learning,kisel2025flaws}. \\
        The results exhibit the phenomenon of drastic drop in accuracy for low values of $\lambda$, for which \ac{nvp} performs the best in Koo-Fu transformed feature spaces.
    }
    \label{tab:zeroshot_clip}
\end{table}

\noindent\textbf{Metrics}
Our primary metric is \emph{accuracy}, that is, the ratio of correctly classified samples.
We are aware that this metric may be overly simplistic since in the event of a misclassification it does not convey the severity of the resulting error, and it may also be sensitive to annotation noise.
Therefore,
%in line with established practice \todo{[cite]},
we additionally report the generalized \emph{top-k accuracy}.
If supported by the classification method, we have the classifier output $k$~samples with the highest activation instead of just one.
If the ground truth class corresponds to any of the retrieved samples, it is considered a match.

Contrary to originally published \imnet[1], the “cleaner” validation subset (see above) assigns multiple labels to some images. In such cases, we adopt the \real{} accuracy \cite{kisel2025flaws} where a decision is considered correct if any of the ground truth labels matches any of the retrieved samples.

\noindent\textbf{Evaluation}
In all experiments, we follow the pipeline outlined in \cref{sec:pipeline}.
Unless otherwise noted, we learn separate Koo-Fu transforms from \imnet[1] (denoted as 1K) and \imnet[21] (denoted as 21K) training samples in CLIP \mbox{ViT-L/14} 768-dimensional feature space and evaluate their accuracy on the validation split(s) of \imnet[1].
The source dataset used to calculate the respective matrix (see equations in \cref{sec:method:prelim}) is denoted by subscript (\eg, $\mathbf{S}_{w:\mathrm{1K}}$).
Analogous experiments that do not use Koo-Fu transform are regarded as the baseline.
All embeddings are normalized to the unit length prior to evaluation, corresponding to using \emph{cosine similarity} as the metric for distance computation.

\subsection{Results}

\noindent\textbf{Vision-only space}
\label{sec:vision}
\begin{table}[t]
    \centering
    \begin{tabular}{crrrrr}
        \toprule
        Space & & \multicolumn{2}{c}{Top-1} & \multicolumn{2}{c}{Top-5} \\
        $f(\cdot)$ & \multicolumn{1}{c}{Dim.} & \multicolumn{1}{c}{Clean} & \multicolumn{1}{c}{Val.} & \multicolumn{1}{c}{Clean} & \multicolumn{1}{c}{Val.} \\
        \midrule
        (--) & 768 & 85.82 & 75.08 & 96.84 & 93.42 \\
        (1K) & 768 & 89.71 & 79.05 & 98.19 & 95.54 \\
        \midrule
        (1K) & 640 & \textbf{89.72} & 79.06 & 98.19 & 95.54 \\
        (1K) & 512 & 89.69 & \textbf{79.07} & \textbf{98.23} & \textbf{95.65} \\
        (1K) & 384 & 89.70 & 79.03 & \textbf{98.23} & \textbf{95.65} \\
        (1K) & 256 & 89.53 & 78.80 & 98.14 & 95.49 \\
        (1K) & 192 & 89.22 & 78.45 & 97.99 & 95.25 \\
        (1K) & 128 & 88.39 & 77.50 & 97.61 & 94.57 \\
        (1K) &  96 & 87.33 & 76.43 & 97.11 & 93.89 \\
        (1K) &  76 & 86.18 & 75.30 & 96.64 & 93.17 \\
        (1K) &  64 & 85.00 & 74.22 & 96.12 & 92.39 \\
        (1K) &  32 & 76.98 & 66.21 & 92.43 & 87.55 \\
        \bottomrule
    \end{tabular}
    \caption{
        {\bf Accuracy of \ac{nvp} in lower-dimensional subspaces.} 
        Measured on {\bf \imnet[1]} validation and ``clean'' subsets.
        The subspaces are obtained by the Fukunaga-Koontz transform using
        $\mathbf{S}'^{}_{w:\mathrm{1K}}$ ($\lambda = 50$) and $\mathbf{S}'^{}_{b:\mathrm{1K}}$.
        Dimensions aligned with eigenvectors with the smallest eigenvalues of $\mathbf{S}'_{b:\mathrm{1K}}$ (\cref{eq:Sb-after-whitening}) are discarded.
        The top two rows provide full space results from \cref{tab:vnp_clip}, rows~1 and 2, for easier comparison.
    }
    \label{tab:dim_red_1k}
\end{table}
\begin{table}[t]
    \setlength{\tabcolsep}{4pt}
    \centering
    \begin{tabular}{crrrrrrr}
        \toprule
        Space & & \multicolumn{2}{c}{1-NN} & \multicolumn{2}{c}{10-NN} & \multicolumn{2}{c}{15-NN} \\
        $f(\cdot)$ & \multicolumn{1}{c}{Dim.} & \multicolumn{1}{c}{Clean} & \multicolumn{1}{c}{Val.} & \multicolumn{1}{c}{Clean} & \multicolumn{1}{c}{Val.} & \multicolumn{1}{c}{Clean} & \multicolumn{1}{c}{Val.} \\
        \midrule
        (--) & 768 & 84.9 & 76.3 & 88.7 & 79.6 & 88.7 & 79.5 \\
        (1K) & 768 & 86.1 & 77.6 & 89.9 & 81.0 & 90.1 & 81.0 \\
        \midrule
        (1K) & 640 & 86.1 & 77.7 & 89.9 & 81.0 & 90.1 & 81.0 \\
        (1K) & 512 & 86.7 & 78.4 & 90.4 & 81.6 & 90.6 & 81.7 \\
        (1K) & 384 & 87.2 & 79.0 & 91.0 & 82.2 & 91.1 & 82.0 \\
        (1K) & 256 & \textbf{87.7} & \textbf{79.4} & \textbf{91.2} & \textbf{82.4} & \textbf{91.3} & \textbf{82.3} \\
        (1K) & 192 & \textbf{87.7} & \textbf{79.4} & 91.1 & 82.2 & 91.2 & 82.2 \\
        (1K) & 128 & 87.3 & 78.8 & 90.8 & 81.8 & 91.0 & 81.7 \\
        (1K) &  96 & 86.9 & 78.1 & 90.3 & 81.1 & 90.2 & 81.0 \\
        (1K) &  76 & 86.2 & 77.2 & 89.7 & 80.5 & 89.6 & 80.2 \\
        (1K) &  64 & 85.5 & 76.4 & 88.9 & 79.5 & 88.9 & 79.4 \\
        (1K) &  32 & 79.0 & 69.1 & 83.3 & 73.3 & 83.3 & 73.1 \\
        \bottomrule
    \end{tabular}
    \caption{
        {\bf Accuracy of \ac{knn} in lower-dimensional subspaces} for $k = 1, 10\ \text{and} \ 15$.
        Setting identical to \cref{tab:dim_red_1k} -- $\mathbf{S}'_{w:\mathrm{1K}}$ computed from samples in CLIP 1K ($\lambda = 150$), evaluated on {\bf \imnet[1]} validation sets, preserving only the most discriminative dimensions from $\mathbf{S}'_{b:\mathrm{21K}}$.
        The top two rows display prior results from \cref{tab:knn_cosine_clip} for easier comparison.
    }
    \label{tab:dimred_knn}
\end{table}

The initial set of experiments deals with classification of samples in vision-only feature spaces, namely with the nearest visual prototype (\ac{nvp}) and the \ac{knn} classifiers.
No dimensionality reductions (\cref{sec:method:prelim}) are performed, all computations are made in a transformed embedding space having the same dimensionality.

The results are reported in \cref{tab:vnp_clip}. We observe that the baseline accuracy of nearest visual prototype classification on \imnet[1] validation dataset performed in the unmodified CLIP feature space is consistently improved by Koo-Fu transform across metrics (rows 1--3).
In particular, the transform learned from \imnet[1] samples increases \real{} (\mbox{top-1}) accuracy from 85.2\% to 89.7\% and from 75.1\% to 79.1\% for the standard evaluation protocol.
Similarly, top-5 accuracy is increased by more than 1\% and 2\% to 98.2\% and 95.5\%, respectively, reducing the number of failures by approximately 40\%.

Slightly smaller improvement is observed when learning Koo-Fu transform from \imnet[21] samples, to 88.1\% and 78.4\% in top-1 and 98.0\% and 95.3\% in top-5, respectively.
As this transform is learned from ten times more samples and twenty times as many classes, its generality is naturally traded off for performance on a particular subset.

We also perform the same experiments with larger sets of classes, namely 14K and 21K (see \cref{sec:setup} §~Datasets and \cref{tab:synset_groups}).
These supersets of \imnet[1] emulate an image retrieval setting in which many \emph{distractor} images are present.
Even under these circumstances, \kf{} repeats previously observed improvements, increasing baseline accuracy by up to 4\% (rows 4--9).

\Cref{tab:knn_cosine_clip} presents results of experiments similar to those in \cref{tab:vnp_clip}, substituting classification by nearest visual prototype with classification by \knn{} (\ac{knn}), reporting only top-1 accuracy, while testing different values of~$k$.
In general, we observe a rather smaller improvement caused by Koo-Fu transform than in the case of \ac{nvp}.
In fact, \ac{knn} already outperforms \ac{nvp} in the unmodified CLIP space by 3--4.5\% when $k =$ 10 or 15, while being on parity when $k = 1$ (rows 1, 4, and 7).
In transformed spaces, \mbox{10-NN} and \mbox{15-NN} outperform \ac{nvp} by 4--6\% in \real{} and standard evaluation and settings with and without distractors.
However, this comes at the cost of classification being $1000\times$ less memory efficient and slower.
%\footnote{The speed difference of finding the nearest among the 1K prototypes and among the 1~million IN-1k training images depends on the nearest neighbor search method used.}.

As noted in \cref{sec:method:prelim}, the whitening transform requires application of regularization to produce meaningful results.
Tables~\ref{tab:vnp_clip} and \ref{tab:knn_cosine_clip} report measurements observed with the parameter $\lambda = 150$ in 1K space and $\lambda = 1500$ in 21K space.
More details on the influence of the parameter $\lambda$ on performance are provided in the ablation study (\cref{sec:ablation:reg}).

\noindent\textbf{Zero-shot classification}
\label{sec:zeroshot}

The next experiments involve classification by textual embeddings, called \emph{zero-shot classification}.
This setting is a modification of \ac{nvp} where visual prototypes are replaced with the corresponding textual prototypes, obtained by encoding class names using the CLIP text encoder.

As noted earlier, the nature of training data for CLIP-like models poses a challenge for optimal evaluation. Therefore, following \cite{kisel2025flaws}, we employ more prompting strategies:

\begin{enumerate}
    \item \textit{synset} -- embedding produced by the prompt template ``image of \{class name\}'', substituting class names from the original WordNet synset \cite{miller1995wordnet},
    \item \textit{synset}$_7$ -- mean of 7 prompt embeddings of templates from \cite{kisel2025flaws}, using the same class names,
    \item \textit{mod}$_7$ -- mean of the same 7 prompt templates, where IN-1k class names are replaced with OpenAI optimized names \cite{radford2021learning}.
\end{enumerate}

The results are shown in \cref{tab:zeroshot_clip}.
%Consistent with prior ...
We observe that the selected prompting strategy has considerable influence on accuracy, with \textit{mod}$_7$ performing the best and maintaining parity with the \ac{nvp} approach (\cref{sec:vision}).
However, the most noticeable phenomenon is that the classification accuracy deteriorates dramatically for low values of the parameter $\lambda$ (\eg, depending on the prompting strategy, down to 1--5\% with $\lambda = 50$), revealing a possible modality gap in our approach.
To maintain zero-shot accuracy from the unmodified space, the parameter $\lambda$ has to be around 5k. However, this leads to a drop in \ac{nvp} accuracy by 1.5\% (discussed later, see \cref{sec:ablation:reg}), exposing a trade-off in the choice of the modality.

Zero-shot accuracy in the regularized space is also strongly affected by dimensionality reduction (more on this in \cref{sec:dimred}).

\noindent\textbf{Dimensionality reduction}
\label{sec:dimred}
\begin{table}[t]
    \setlength{\tabcolsep}{3pt}
    \centering
    \begin{tabular}{crrrrrrr}
        \toprule
        Space & & \multicolumn{2}{c}{\textit{synset}} & \multicolumn{2}{c}{\textit{synset}$_7$} & \multicolumn{2}{c}{\textit{mod}$_7$} \\
        $f(\cdot)$ & \multicolumn{1}{c}{Dim.} & \multicolumn{1}{c}{Clean} & \multicolumn{1}{c}{Val.} & \multicolumn{1}{c}{Clean} & \multicolumn{1}{c}{Val.} & \multicolumn{1}{c}{Clean} & \multicolumn{1}{c}{Val.} \\
        \midrule
        (--) & 768 & 79.60 & 69.79 & 83.42 & 73.50 & 85.53 & 75.38 \\
        (1K) & 768 & 11.93 & 10.06 & 26.61 & 22.34 & 45.34 & 38.55 \\
        \midrule
        (1K) & 640 &  7.38 &  6.17 & 21.69 & 18.10 & 44.15 & 37.53 \\
        (1K) & 512 & 72.37 & 77.11 & 79.03 & 68.99 & 81.27 & 71.11 \\
        (1K) & 384 & \textbf{77.11} & \textbf{67.05} & \textbf{80.46} & \textbf{70.25} & \textbf{82.30} & \textbf{72.01} \\
        (1K) & 256 & 76.07 & 66.11 & 79.60 & 69.39 & 81.52 & 71.19 \\
        (1K) & 192 & 75.33 & 65.28 & 78.50 & 68.27 & 80.69 & 70.35 \\
        (1K) & 128 & 70.51 & 61.02 & 74.28 & 64.49 & 76.98 & 66.82 \\
        (1K) &  96 & 66.17 & 56.90 & 70.31 & 60.69 & 72.92 & 63.02 \\
        (1K) &  76 & 60.99 & 52.28 & 65.79 & 56.80 & 68.00 & 58.82 \\
        (1K) &  64 & 55.80 & 47.93 & 61.51 & 52.91 & 63.80 & 54.96 \\
        (1K) &  32 & 29.81 & 25.63 & 33.94 & 29.31 & 36.17 & 31.35 \\
        \bottomrule
    \end{tabular}
    \caption{
        {\bf Top-1 accuracy of zero-shot classification in lower-dimensional subspaces.}
        Measured on {\bf \imnet[1]} validation and ``clean'', with $\mathbf{S}'_{w:\mathrm{1K}}$ regularized by $\lambda = 150$, dropping dimensions aligned with eigenvectors of $\mathbf{S}'_{b:\mathrm{1K}}$ with the smallest eigenvalues.
        The top two rows repeat results from \cref{tab:zeroshot_clip} for reference.
    }
    \label{tab:dim_reduction_zeroshot}
\end{table}
\begin{table}[t]
    \setlength{\tabcolsep}{3pt}
    \centering
    \begin{tabular}{cr cr r r r}
        \toprule
        Space & \multicolumn{1}{c}{\multirow{2}{*}{Dim.}} & \multicolumn{2}{c}{15-NN} & \multicolumn{1}{c}{Search} & \multicolumn{1}{c}{\multirow{2}{*}{Speedup}} & \multicolumn{1}{c}{Memory} \\
        $f(\cdot)$ & & \multicolumn{1}{c}{Clean} & \multicolumn{1}{c}{Val.} & \multicolumn{1}{c}{time} & & \multicolumn{1}{c}{(MB)} \\
        \midrule
        (--) & 768 &       88.7 &       79.5 & \multirow{2}{*}{2.47s} & \multirow{2}{*}{---} & \multirow{2}{*}{3753.4} \\
        (1K) & 768 &       90.1 &       81.0 & & & \\
        \midrule
        (1K) & 640 &       90.1 &       81.0 &     2.37s &   1.04× &   3127.8 \\
        (1K) & 512 &       90.6 &       81.7 &     1.97s &   1.25× &   2502.3 \\
        (1K) & 384 &       91.1 &       82.0 &     1.57s &   1.57× &   1876.7 \\
        (1K) & 256 & \textbf{91.3} & \textbf{82.3} & 1.23s & 2.01× &   1251.1 \\
        (1K) & 192 &       91.2 &       82.2 &     1.00s &   2.47× &    938.4 \\
        (1K) & 128 &       91.0 &       81.7 &     0.76s &   3.24× &    625.6 \\
        (1K) &  96 &       90.2 &       81.0 &     0.63s &   3.92× &    469.2 \\
        (1K) &  76 &       89.6 &       80.2 &     0.59s &   4.20× &    371.4 \\
        (1K) &  64 &       88.9 &       79.4 &     0.55s &   4.52× &    312.8 \\
        (1K) &  32 &       83.3 &       73.1 &     0.45s &   5.49× &    156.4 \\
        %(1K) & 16 &       69.2 &       59.2 &     0.41s &   6.03× &     78.2 \\
        %(1K) &  8 &       40.8 &       34.3 &     0.40s &   6.20× &     39.1 \\
        \bottomrule
    \end{tabular}

    \caption{
        {\bf Efficiency and resource consumption of 15-NN classification.}
        Precision data correspond to \cref{tab:dimred_knn}.
        %Evaluated on the IN-1k Validation set using Koo-Fu transformed 1K features ($\lambda = 150$).
        {\bf Search Time} and {\bf Memory} are measured on an H200 nVidia GPU while using FAISS \cite{douze2024faiss}. \\
        With 256-dimensional projection, 15-NN achieves improved accuracy while delivering 2× speedup.
        64-dimensional embeddings preserve baseline accuracy and increase efficiency 4.5×.
    }
    \label{tab:dimred_15nn}
\end{table}

The last in the series of main experiments deals with dimensionality reduction by \kf{} in addition to the whitening transform, focusing on a potential efficiency--accuracy trade-off.

\Cref{tab:dim_red_1k} presents results of an experiment in which we repeatedly apply the Koo-Fu transform learned from CLIP 1K ($\lambda = 150$) and carry out \ac{nvp} classification on \imnet[1] while gradually decreasing the number of dimensions remaining in the target embedding space.
We observe that the accuracy drops only negligibly (less than 0.5\%) up to 256 dimensions (\ie, 3× reduction), with a very slight increase when dropping only 20--30\% of the least significant directions.
Embeddings with 256 to 76 dimensions provide efficiency--accuracy trade-offs.
Note that the top-1 accuracy in 76 dimensions of the Fukunaga-Koontz transformed space is still superior to the original 10× larger CLIP space.

%\todo{Possibly include \cref{tab:dim_red_21k}, too, and explain its evaluation protocol.}

\Cref{tab:dimred_knn} shows results from an analogous experiment where we perform classification by \ac{knn}.
The observations are even more favorable, with the threefold reduction (256~dim.) increasing accuracy by 1\%, the 10-fold reduction (76~dim.) being on par with no reduction made and the 12-fold reduction (64~dim.) comparable to baseline accuracy when whitening is not involved at all.
This demonstrates multiple benefits of dimensionality reduction by \kf{}, permitting both efficiency and accuracy boost or a trade-off between them.
The influence of dimensionality reduction on other aspects of \ac{knn} is discussed in the ablation study (\cref{sec:ablation:dimred}).

Finally, \Cref{tab:dim_reduction_zeroshot} presents results for when dimensionality reduction of \kf{} is applied to zero-shot classification.
As demonstrated in \cref{sec:zeroshot}, performance is severely harmed when switching from visual to textual class prototypes without further increasing the regularization factor.
Surprisingly, dimensionality reduction causes the accuracy for low $\lambda$ to almost recover, where 50\% reduction (384~dim.) underperforms baseline accuracy in unmodified CLIP space only by 3--4\%. We do not have a precise explanation for this phenomenon.
%\todo{but?}

\subsection{Ablation study}
\label{sec:ablation}

\noindent\textbf{Regularization}
\label{sec:ablation:reg}
\begin{table}[t]
    \centering
    \begin{tabular}{crrrrr}
        \toprule
        Space & & \multicolumn{2}{c}{Top-1} & \multicolumn{2}{c}{Top-5} \\
        $f(\cdot)$ & \multicolumn{1}{c}{$\lambda$} & \multicolumn{1}{c}{Clean} & \multicolumn{1}{c}{Val.} & \multicolumn{1}{c}{Clean} & \multicolumn{1}{c}{Val.} \\
        \midrule
        (--) &  --- & 85.82 & 75.08 & 96.85 & 93.42 \\
        \midrule
        (1K) & \textless{}28 & --- & --- & --- & --- \\
        (1K) &    30 & 89.66 & 78.97 & 98.10 & 95.38 \\
        (1K) &    50 & 89.65 & 78.98 & 98.14 & 95.43 \\
        (1K) &   100 & 89.69 & 79.04 & 98.18 & 95.52 \\
        (1K) &   150 & \textbf{89.71} & \textbf{79.05} & 98.19 & 95.54 \\
        (1K) &   200 & 89.69 & 79.02 & \textbf{98.20} & \textbf{95.55} \\
        (1K) &   250 & 89.67 & 79.03 & 98.19 & 95.54 \\
        (1K) &   500 & 89.58 & 78.94 & 98.16 & 95.53 \\
        (1K) &  1000 & 89.39 & 78.67 & 98.09 & 95.39 \\
        %(1K) &  3000 & 88.59 & 77.96 & 97.93 & 95.06 \\
        (1K) &  5000 & 88.15 & 77.50 & 97.77 & 94.85 \\
        (1K) & 10000 & 87.43 & 76.79 & 97.54 & 94.45 \\
        \bottomrule
    \end{tabular}
    \caption{
        The impact of $\mathbf{S}'_{w:\mathrm{1K}}$ regularization on IN-1k \ac{nvp} accuracy in the whitened CLIP space.
        $\mathbf{S}_{w:\mathrm{1K}}$ is accumulated from more than 1.2~million IN-1k training images, and $\mathbf{S}_{w:\mathrm{1K}}$ elements on the diagonal range from 76 to 4318.\\
        For $\lambda <28$, the scipy \texttt{eigh()} procedure outputs, incorrectly (the matrix is constructed as symmetric positive-semidefinite), also negative eigenvalues. Once the problem is addressed by regularization, the value of $\lambda$ has minimal impact on the accuracy.
    }
    \label{tab:abl:reg_in1kt}
\end{table}

\begin{table}[t]
    \centering
    \begin{tabular}{crrrrr}
        \toprule
        Space & & \multicolumn{2}{c}{Top-1} & \multicolumn{2}{c}{Top-5} \\
        $f(\cdot)$ & \multicolumn{1}{c}{$\lambda$} & \multicolumn{1}{c}{Clean} & \multicolumn{1}{c}{Val.} & \multicolumn{1}{c}{Clean} & \multicolumn{1}{c}{Val.} \\
        \midrule
         (--) &   --- & 85.82 & 75.08 & 96.85 & 93.42 \\
        \midrule
        (21K) & \textless{215} & --- & --- & --- & --- \\
        (21K) &   250 & 88.97 & 78.22 & 97.92 & 95.14 \\
        (21K) &   500 & 89.00 & 78.29 & 97.96 & 95.19 \\
        (21K) &   750 & 89.03 & 78.35 & 97.98 & 95.23 \\
        (21K) &  1000 & 89.06 & 78.37 & 98.01 & 95.27 \\
        %(21K) &  1250 & 89.08 & 78.38 & 98.03 & 95.29 \\
        (21K) &  1500 & \textbf{89.11} & \textbf{78.40} & 98.03 & 95.31 \\
        (21K) &  2000 & 89.08 & 78.39 & \textbf{98.06} & 95.34 \\
        (21K) &  2500 & 89.08 & 78.39 & \textbf{98.06} & \textbf{95.35} \\
        (21K) &  3000 & 89.09 & 78.39 & 98.05 & 95.33 \\
        (21K) &  5000 & 89.06 & 78.34 & 98.04 & 95.31 \\
        (21K) & 10000 & 88.89 & 78.17 & 97.98 & 95.18 \\
        \bottomrule
    \end{tabular}
    \caption{
    The impact of $\mathbf{S}'_{w:\mathrm{21K}}$ regularization on IN-1k \ac{nvp} accuracy in the whitened CLIP space.
    The matrix is accumulated from more than 13~million IN-21k images.
    Diagonal elements of $\mathbf{S}_{w:\mathrm{21K}}$ range from 924 to 39,291. \\
    For $\lambda < 215$, the scipy \texttt{eigh()} procedure outputs also negative eigenvalues.
    In experiments, $\mathbf{S}_{w:\mathrm{21K}}$ is not used as performance with $\mathbf{S}_{w:\mathrm{1K}}$ is slightly better.
    }
    \label{tab:abl:reg_in21kt}
\end{table}

We discuss the influence of the $\lambda$ regularization parameter on performance of \kf{}.
We start by finding the lowest $\lambda$, for which all eigenvalues of $\mathbf{S}'_w$ (\cref{eq:S_w_reg}) are positive.
Then, we continue by gradually increasing its value and measuring the accuracy.

The results are presented in Tables~\ref{tab:abl:reg_in1kt} and \ref{tab:abl:reg_in21kt}.
We validate on \imnet[1] in 1K and 21K CLIP spaces, respectively.
We observe that the lower bound already provides a reasonable estimate and loses only about 0.1\% in accuracy relative to the best-performing value ($\lambda = 150$ for 1K and $\lambda = 1500$ for 21K).

In general, the impact of selecting a particular value for $\lambda$ is minimal. The accuracy decreases significantly only for relatively high values of $\lambda$.

\noindent\textbf{\textit{k}-NN dimensionality reduction}
\label{sec:ablation:dimred}

We also discuss influence of \kf{} dimensionality reduction on practical aspects of \ac{knn}.
Our empirical measurements are reported in \cref{tab:dimred_15nn}.
In particular, applying dimensionality reduction optimized for maximal accuracy yields a twofold speedup, whereas a more aggressive reduction that preserves baseline accuracy decreases runtime to 22\% (with memory demands reduced proportionally to the reduction factor).

\noindent\textbf{Nearest neighbor count selection}
\label{sec:ablation:nn}

Prior to experiments with \ac{knn}, we performed a search for the optimal number of nearest neighbors.
Depending on the setting, the value is located in range 10--17; detailed results are available in the supplementary material.%(\cref{tab:knn_clip_results}).
% We investigate the sensitivity of the \ac{knn} classifier to the neighbor count $k$ in \cref{tab:knn_clip_results}.
% While using a single neighbor ($k=1$) yields the lowest precision, increasing $k$ provides a consistent performance gain that stabilizes between $k=10$ and $k=17$. 
% Notably, \kf{} consistently outperforms the baseline by more than 1\% across all values of $k$.
% We fix $k=15$ for subsequent experiments as a robust balance between precision and computational cost.

\noindent\textbf{Embedding normalization}
\label{sec:ablation:distance}

In addition to the main experiments, we test the behavior of \kf{} when we relax our evaluation protocols and omit embedding normalization, selecting the best candidate in the feature space based on the Euclidean distance.

Our observations indicate that raw embeddings almost achieve performance parity with normalized ones, leaving a gap of \textless{1\%} in most settings.
This further underscores the role of the whitening transformation in our pipeline as a regularization of the feature space.
However, we also note the need to set the parameter $\lambda$ to very high values (tens or hundreds of thousands).

The results are provided in the supplementary material.
\section{Conclusions}
\label{sec:concl}
We presented \kf{}, a simple yet effective closed-form adaptation method for vision–language models based on applying a regularized Fukunaga-Koontz (whitened Linear Discriminant Analysis) transform directly to frozen VLM embeddings.
We demonstrate the method on CLIP, a canonical VLM.
Across different classification setups, this leads to consistent improvements in performance and clear trade-offs between accuracy, dimensionality, and inference cost.
These results highlight representation-space adaptation as a promising and underexplored direction for leveraging foundation models.
Future work will focus on a deeper theoretical analysis of the transform and its extension to other vision–language and multimodal foundation models.

\clearpage
% Place the bibliography (from references.bib) at the end:
{
    \small
    \bibliographystyle{ieeenat_fullname}
    \bibliography{references}
}

\clearpage
\appendix
\section{Additional experiments}

\begin{table}[h]
    \setlength{\tabcolsep}{5.5pt}
    \centering
    \begin{tabular}{rrrrrrr}
        \toprule
        & \multicolumn{2}{c}{(--)} & \multicolumn{2}{c}{(1K)} & \multicolumn{2}{c}{(21K)} \\
        \multicolumn{1}{c}{$k$} & \multicolumn{1}{c}{Clean} & \multicolumn{1}{c}{Val.} & \multicolumn{1}{c}{Clean} & \multicolumn{1}{c}{Val.} & \multicolumn{1}{c}{Clean} & \multicolumn{1}{c}{Val.} \\
        \midrule
          1 & 84.89 & 76.31 & 86.07 & 77.63 & 85.73 & 77.31 \\
          % 2 & 84.89 & 76.31 & 86.07 & 77.63 \
          % 3 & 86.91 & 78.23 & 88.32 & 79.82 & 87.94 & 79.38 \\
          % 4 & 87.51 & 78.79 & 88.81 & 80.29 \
          5 & 88.01 & 79.22 & 89.25 & 80.62 & 88.94 & 80.22 \\
          % 7 & 88.37 & 79.53 & 89.59 & 80.85 & 89.28 & 80.55 \\
          % 8 & 88.52 & 79.56 & 89.78 & 80.97 \
          9 & 88.67 & 79.63 & 89.84 & 80.97 & 89.52 & 80.64 \\
          10 & 88.73 & \textbf{79.64} & 89.88 & 81.00 & 89.58 & 80.64 \\
          11 & \textbf{88.74} & 79.62 & 89.96 & 80.97 & 89.61 & 80.67 \\
          13 & 88.66 & 79.57 & 90.06 & \textbf{81.01} & 89.69 & \textbf{80.72} \\
          15 & 88.69 & 79.45 & 90.07 & 80.95 & \textbf{89.77} & 80.70 \\
          16 & 88.68 & 79.39 & \textbf{90.09} & 80.93 & 89.71 & 80.69 \\
          17 & 88.66 & 79.36 & \textbf{90.09} & 80.92 & 89.72 & 80.61 \\
          19 & 88.67 & 79.22 & 90.07 & 80.84 & 89.72 & 80.54 \\
          32 & 88.51 & 78.76 & 90.02 & 80.28 & 89.70 & 80.06 \\
          64 & 88.04 & 77.89 & 89.63 & 79.61 & 89.39 & 79.32 \\
          % 128 & 87.19 & 76.72 & 89.10 & 78.69 \
        \bottomrule
    \end{tabular}
    \caption{
        Ablation of \ac{knn} accuracy on \imnet[1] across different neighbor counts ($k$) in untransformed space and whitened CLIP spaces.
        % comparing baseline (--) against Koo-Fu transformed spaces using 1K and 21K statistics. 
        While using a single neighbor ($k=1$) yields the lowest precision, 
        % A clear trend emerges: using a single neighbor ($k=1$) yields the lowest accuracy. % likely due to sensitivity to outliers or label noise in the reference database. 
        increasing $k$ provides a consistent performance gain that stabilizes between $k=10$ and $k=17$. 
        % Specifically, the untransformed space peaks at $k=10$ (79.64\% Val.), while 1K and 21K spaces reach maximums at $k=13$ (81.01\% and 80.72\% Val.). 
        Notably, \kf{} consistently outperforms the baseline by more than 1\% across all values of $k$.
        % and remains remarkably stable within the $k \in [9, 19]$ range, suggesting low sensitivity to this hyperparameter. 
        We fix $k=15$ for subsequent experiments as a balance between precision and computational cost.
    }
    \label{tab:knn_clip_results}
\end{table}

\begin{table}[h]
\setlength{\tabcolsep}{4pt}
    \centering
    \begin{tabular}{F c r r c cccc}
        \toprule
        \# &  & Space & & \multicolumn{2}{c}{Top-1} & \multicolumn{2}{c}{Top-5} \\
        & \multicolumn{1}{c}{} &
        \multicolumn{1}{c}{$f(\cdot)$} &
        \multicolumn{1}{c}{class.} &
        \multicolumn{1}{c}{Clean} &
        \multicolumn{1}{c}{Val.} &
        \multicolumn{1}{c}{Clean} &
        \multicolumn{1}{c}{Val.} \\
        \midrule
        \textcolor{gray}{1} & \multirow{9}{*}{\rotatebox{90}{}}
          & (--)  & \multirow{3}{*}{1K}  & 85.37 & 74.58 & 96.67 & 93.12 \\
        \textcolor{gray}{2} &
          & (1K)  &  & \textbf{89.58} & \textbf{78.92} & \textbf{98.10} & \textbf{95.34} \\
        \textcolor{gray}{3} &
          & (21K) &  & 88.91 & 78.18 & 97.84 & 95.02 \\
        \cmidrule{3-8}
        \textcolor{gray}{4} &
          & (--)  & \multirow{3}{*}{14K}  & 65.44 & 57.47 & 73.16 & 69.90 \\
        \textcolor{gray}{5} &
          & (1K)  &  & \textbf{68.22} & \textbf{60.44} & \textbf{74.08} & \textbf{71.22} \\
        \textcolor{gray}{6} &
          & (21K) &  & 67.78 & 59.88 & 73.95 & 71.01 \\
        \cmidrule{3-8}
        \textcolor{gray}{7} &
          & (--)  & \multirow{3}{*}{21K}  & 50.57 & 42.34 & 79.49 & 70.64 \\
        \textcolor{gray}{8} &
          & (1K)  &  & \textbf{55.12} & \textbf{46.63} & \textbf{83.87} & \textbf{75.31} \\
        \textcolor{gray}{9} &
          & (21K) &  & 53.48 & 45.08 & 83.04 & 74.32 \\
        \bottomrule
    \end{tabular}
    \caption{
        {\bf Accuracy of \ac{nvp} classification across different tasks} (Euclidean distance).
        We compare baseline CLIP embeddings (--) against spaces transformed with 1K and 21K statistics. 
        Notably, the 1K space consistently delivers the highest accuracy across all benchmarks, 
        % outperforming the (21K) space even on 14K and 21K classification tasks. 
        improving the Val. baseline in \mbox{top-1} and \mbox{top-5} by more than 4\%.
        % This suggests that 1K-based statistics provide a more robust and generalizable feature space.
        Results use $\lambda = 50\mathrm{k}$ for (1K) and $\lambda = 100\mathrm{k}$ for (21K).
    }
    \label{tab:centroids_l2_clip}
\end{table}
\begin{table}[tb]
    \centering
    \begin{tabular}{crrrrr}
        \toprule
        Space & & \multicolumn{2}{c}{Top-1} & \multicolumn{2}{c}{Top-5} \\
        $f(\cdot)$ & \multicolumn{1}{c}{Dim.} & \multicolumn{1}{c}{Clean} & \multicolumn{1}{c}{Val.} & \multicolumn{1}{c}{Clean} & \multicolumn{1}{c}{Val.} \\
        \midrule
        (--) & 768 & 85.37 & 74.58 & 96.67 & 93.12 \\
        (1K) & 768 & 89.03 & 78.16 & 97.94 & \textbf{95.10} \\
        \midrule
        (1K) & 640 & 89.05 & 78.20 & \textbf{97.93} & \textbf{95.10} \\
        (1K) & 512 & \textbf{89.08} & \textbf{78.30} & 97.92 & 95.09 \\
        (1K) & 384 & 89.01 & 78.23 & 97.90 & 95.07 \\
        (1K) & 256 & 88.79 & 77.98 & 97.81 & 94.97 \\
        (1K) & 192 & 88.60 & 77.71 & 97.73 & 94.78 \\
        (1K) & 128 & 87.63 & 76.73 & 97.56 & 94.28 \\
        (1K) &  96 & 86.52 & 75.54 & 97.03 & 93.59 \\
        (1K) &  76 & 85.27 & 74.29 & 96.61 & 92.98 \\
        (1K) &  64 & 84.05 & 73.00 & 96.18 & 92.30 \\
        (1K) &  32 & 75.16 & 64.12 & 92.23 & 86.90 \\
        % (1K) &  16 & 58.57 & 49.13 & 83.02 & 76.32 \\
        % (1K) &   8 & 34.19 & 28.47 & 62.64 & 55.77 \\
        \bottomrule
    \end{tabular}
    \caption{
        {\bf Performance of \ac{nvp} classification under dimensionality reduction.} 
        Results use Euclidean distance in spaces transformed with $\mathbf{S}'_{w:\mathrm{1K}}$ ($\lambda = 50000$). 
        Notably, \kf{} outperforms the 768-dimensional baseline (--) even when reduced to 76 dimensions. 
        % while the peak accuracy is achieved at 512 dimensions (78.30\% Val.). 
        The transformation maintains Top-5 accuracy above 95\% for most configurations.
        % indicating high semantic stability.
        % demonstrating that significant compression can be achieved with minimal loss in precision compared to the full-dimensional whitened space.
    }
    \label{tab:dim_red_1k_centroids_l2}
\end{table}

\begin{table}[t]
    \centering
    \begin{tabular}{crrrrr}
        \toprule
        Space & & \multicolumn{2}{c}{Top-1} & \multicolumn{2}{c}{Top-5} \\
        $f(\cdot)$ & \multicolumn{1}{c}{$\lambda$} & \multicolumn{1}{c}{Clean} & \multicolumn{1}{c}{Val.} & \multicolumn{1}{c}{Clean} & \multicolumn{1}{c}{Val.} \\
        \midrule
        (--) &  --- & 85.37 & 74.58 & 96.67 & 93.12 \\
        \midrule
        %\textless{9343} & --- & --- \\
        (1K) & \textless{9400} & --- & --- & --- & --- \\
        (1K) &  10k & 89.51 & 78.80 & 98.04 & 95.26 \\
        %(1K) &12.5k & 89.55 & 78.84 & 98.05 & 95.28 \\
        %(1K) &  15k & 89.55 & 78.84 & 98.06 & 95.30 \\
        %(1K) &  20k & 89.56 & 78.87 & 98.06 & 95.33 \\
        (1K) &  25k & 89.57 & 78.88 & 98.06 & 95.33 \\
        (1K) &  50k & \textbf{89.58} & \textbf{78.92} & \textbf{98.10} & 95.34 \\
        (1K) &  75k & 89.53 & 78.90 & \textbf{98.10} & 95.34 \\
        (1K) & 100k & 89.49 & 78.88 & \textbf{98.10} & \textbf{95.37} \\
        (1K) & 150k & 89.44 & 78.83 & 98.08 & 95.34 \\
        (1K) & 200k & 89.39 & 78.76 & 98.05 & 95.32 \\
        \bottomrule
    \end{tabular}
    \caption{
    {\bf Regularization ablation for 1K-stats transformation.} 
    We report accuracy as a function of $\lambda$ using Euclidean distance. 
    Performance is highly stable for $\lambda \geq 10\mathrm{k}$, with a peak around 50k. 
    % The baseline (--) is outperformed by over 4\% in Val. accuracy. 
    Values below 9.4k result in an ill-conditioned covariance matrix. 
    % highlighting the necessity of the regularization described in \cref{sec:ablation:reg}.
}
\end{table}
%\input{tables/ablation_l2_21k}

% \clearpage
% \input{sections/tmp}

\end{document}